\documentclass[final]{cvpr}

\usepackage{times}
\usepackage{epsfig}
\usepackage{graphicx}
\usepackage{amsmath}
\usepackage{amssymb}
\usepackage{bm}
\usepackage{enumerate}
\usepackage{enumitem}
\DeclareMathOperator*{\argmax}{arg\,max}

\usepackage{color}
\definecolor{mypink}{RGB}{255,135,180}
\definecolor{myblue}{RGB}{0,0,255}
\definecolor{myyellow}{RGB}{255,200,0}
\definecolor{mygreen}{RGB}{0,100,0}
\definecolor{myorange}{RGB}{255,165,0}
\definecolor{mygold}{RGB}{255,185,0}
\definecolor{mymaroon}{RGB}{128,0,0}
\definecolor{myaquamarine}{RGB}{120,200,190}
\definecolor{myturquoise}{RGB}{64,224,208}
\definecolor{mymagenta}{RGB}{255,0,255}
\definecolor{myviolet}{RGB}{238,130,238}

\usepackage[pagebackref=true,breaklinks=true,colorlinks,bookmarks=false]{hyperref}
\def\cvprPaperID{2351} % *** Enter the CVPR Paper ID here
\def\confYear{CVPR 2021}

\begin{document}

%%%%%%%%% TITLE
\title{Anchor-Constrained Viterbi for Set-Supervised Action Segmentation}

\author{Jun Li\\
Oregon State University\\
{\tt\small liju2@oregonstate.edu}
\and
Sinisa Todorovic\\
Oregon State University\\
{\tt\small sinisa@oregonstate.edu}
}

\maketitle
\thispagestyle{empty}

%%%%%%%%% ABSTRACT
\begin{abstract}
   This paper is about action segmentation under weak supervision in training, where the ground truth provides only a set of actions present, but neither their temporal ordering nor when they occur in a training video. We use a Hidden Markov Model (HMM) grounded on a multilayer perceptron (MLP) to label video frames, and thus generate a pseudo-ground truth for the subsequent pseudo-supervised training. In testing, a Monte Carlo sampling of action sets seen in training is used to generate candidate temporal sequences of actions, and select the maximum posterior sequence. Our key contribution is a new anchor-constrained Viterbi algorithm (ACV) for generating the pseudo-ground truth, where anchors are salient action parts estimated for each action from a given ground-truth set. Our evaluation on the tasks of action segmentation and alignment on the benchmark Breakfast, MPII Cooking2, Hollywood Extended datasets demonstrates our superior performance relative to that of prior work. 
\end{abstract}

%%%%%%%%% BODY TEXT
\section{Introduction}
This paper is about action segmentation by labeling video frames with action classes under weak, set-level supervision in training. In the set-supervised training, the ground truth is a set of actions present in a training video, but their temporal ordering, the number of action instances, and their start and end frames are unknown. This is an important problem arising in many recent applications, such as, those dealing with big video datasets with automatically generated set-level annotations (\eg, via word-based video retrievals from Youtube), for which human annotations are not available (\eg because scaling human annotations over numerous video retrieval results is difficult). Our main challenge is that the provided ground truth -- being a set with arbitrarily ordered distinct labels of action classes -- does not provide sufficient constraints for a reliable learning of action segmentation.
 
Prior work typically adopts the following framework. In training, an action model is used to label every video frame, and thus generate a pseudo-ground truth for the subsequent pseudo-supervised training of the model. In testing, a Monte Carlo sampling is first used to generate candidate temporal sequences of actions, and then the learned model is applied to identify the best scoring candidate sequence as the solution of action segmentation. Differences among existing approaches mostly lie in how the pseudo-ground truth is generated. For example, in  \cite{richard2018action},  binary classifiers are independently trained for each action to label frames using multi-instance learning. Such a training, however, cannot learn temporal spans of and transitions between actions, so these parameters are  heuristically set in \cite{richard2018action} for an HMM inference on test videos. The approach in \cite{li2020set} casts the set-supervised training as the NP-hard ``all-color shortest path" problem \cite{DBLP:journals/corr/BilgeCGSAE15}, where frame labeling of a training video is constrained such that every action label from the ground-truth set appears at least once. For generating the pseudo-ground truth, they infer an HMM using a greedy two-step algorithm called set-constrained Viterbi (SCV). The SCV first produces an initial action segmentation by running the vanilla Viterbi, and then flips low-scoring frame labels to actions that have been missed in the initial segmentation but are present in the ground-truth set. Thus, the SCV produces an approximate solution to the NP-hard problem by projection to the legal domain. There are many shortcomings of \cite{li2020set} related to the heuristic, greedy flipping of frame labels.

% \begin{figure}[t]
% \begin{center}
% \begin{minipage}{0.5\linewidth}
% \centering
% \includegraphics[width=0.95\linewidth]{all-color.pdf}

% (a)
% \end{minipage}\hfill%
% \begin{minipage}{0.5\linewidth}
% \centering
% \includegraphics[width=0.95\linewidth]{Overview.pdf}
   
% (b)
% \end{minipage}
% \caption{(a) An action segmentation can be viewed as a directed path in a densely connected video graph whose nodes are frames and colored edges represent action segments. We cast the problem of generating frame-wise pseudo-ground truth as finding an optimal valid ``all-color shortest path"    \cite{DBLP:journals/corr/BilgeCGSAE15} in which all actions from the ground-truth set occur at least once. The figure shows that a video in general has numerous ``all-color shortest paths", and thus our main challenge is to efficiently infer a globally optimal valid path. (b) For a given training video, we use an HMM grounded on an MLP neural network to find the MAP ``all-color shortest path" frame labeling, which is taken as frame-wise pseudo-ground truth. Our contributions include: (1) Anchor-Constrained Viterbi (ACV) algorithm for estimating the MAP pseudo-ground truth, and (2) Diversity regularization for minimizing correlations of saliency scores of distinct actions along the video. The pseudo-ground truth is used for computing the cross-entropy loss, and thus for pseudo fully-supervised training of our HMM and MLP.}
% \label{fig:Overview}
% \end{center}
% \end{figure}

\begin{figure}[ht]
\begin{center}
\includegraphics[width=0.95\linewidth]{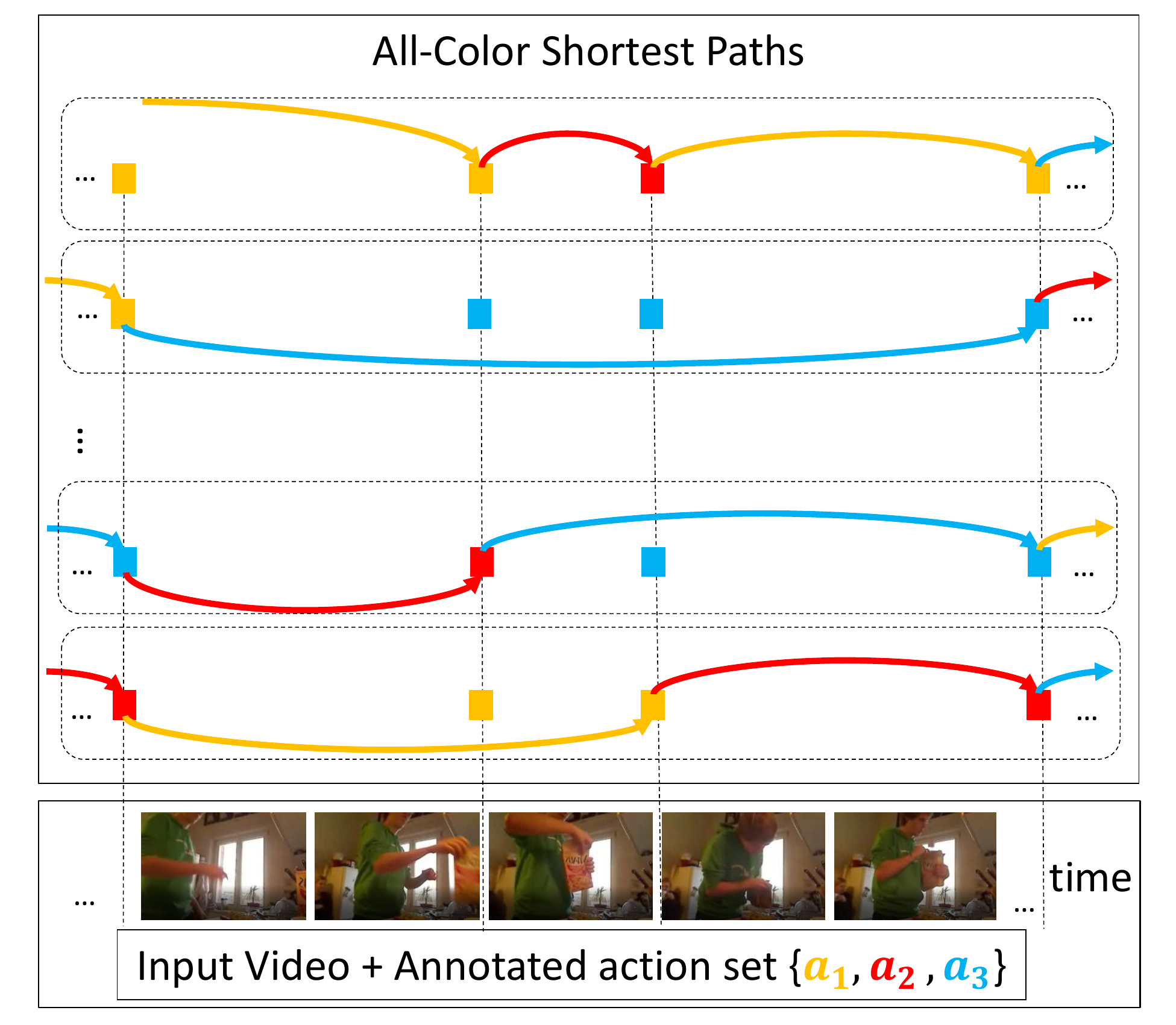}
\end{center}
  \caption{We cast the problem of generating pseudo-ground truth as finding an optimal ``all-color shortest path"    \cite{DBLP:journals/corr/BilgeCGSAE15} in which every action from the ground-truth set occurs at least once. A video has numerous ``all-color shortest paths". Our key contribution is an efficient approximation to this NP-hard problem.}
\label{fig:all-color}
\end{figure}

In this paper, we improve the most critical step of the above framework --  generation of the pseudo-ground truth. As in \cite{li2020set}, we also pose the set-supervised training as  the ``all-color shortest path" problem \cite{DBLP:journals/corr/BilgeCGSAE15}. Our key difference is that we formulate a more effective differentiable approximation to this NP-hard problem that allows for an end-to-end training, reduces training complexity, and ultimately leads to significant performance gains over \cite{li2020set}. 

\begin{figure}[ht]
\begin{center}
\includegraphics[width=0.95\linewidth]{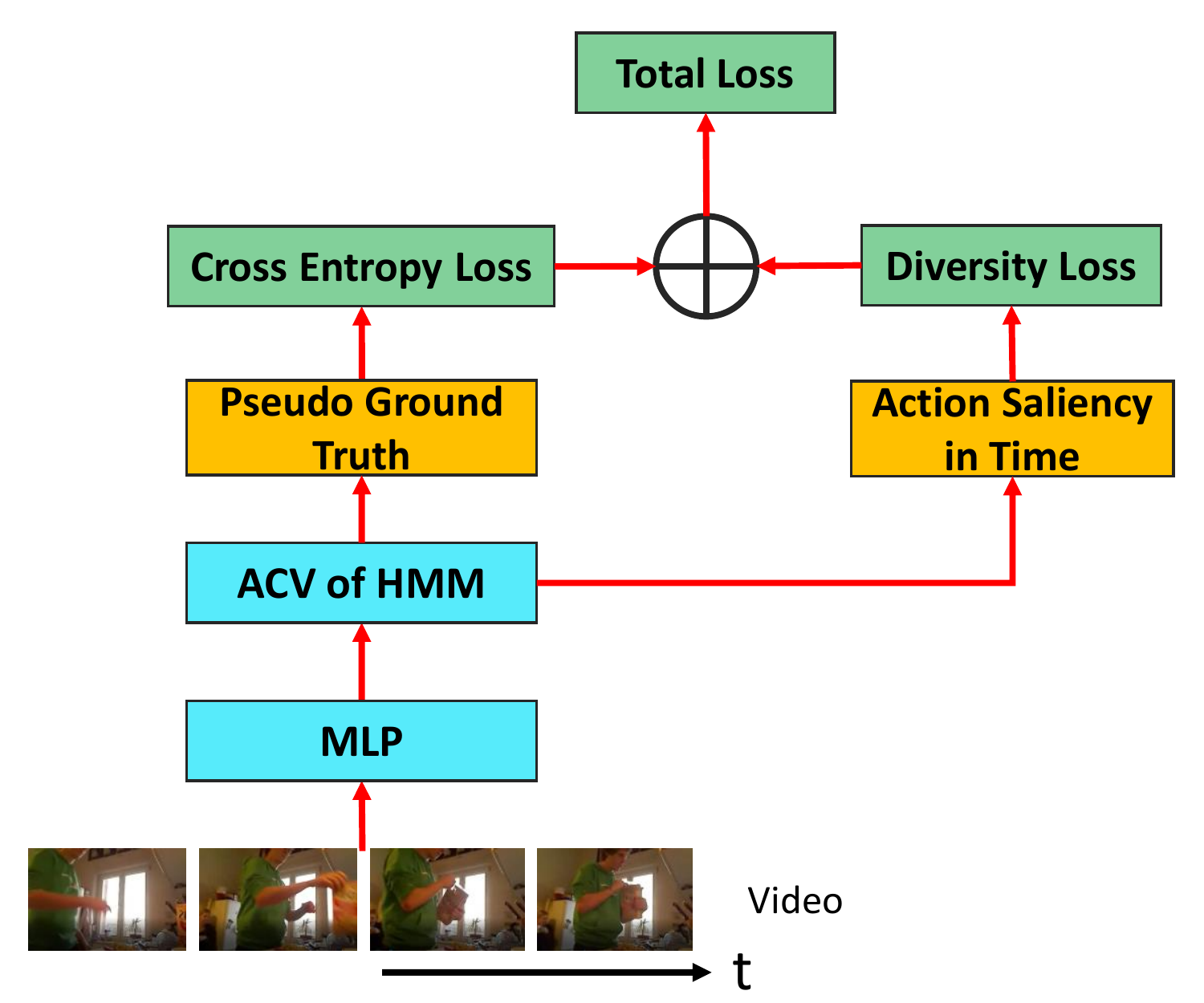}
\end{center}
  \caption{We use an HMM to find the MAP ``all-color shortest path" frame labeling of a training video, which is then taken as the pseudo-ground truth. For the MAP inference in training, we specify the new Anchor-Constrained Viterbi (ACV) algorithm, and diversity regularization which minimizes correlations of detected actions along the video. The pseudo-ground truth is used for computing the cross-entropy loss and regularization in the subsequent fully-supervised training of our HMM and MLP, enabling end-to-end training.}
\label{fig:overview}
\end{figure}

As illustrated in Fig.~\ref{fig:all-color}, all legal action segmentations of a training video can be represented by distinct paths in a directed video segmentation graph. A legal (or valid) path includes every action from the ground-truth set at least once. Our goal is to efficiently identify the highest-scoring valid path. For this scoring, we use an HMM grounded via a two-layer MLP onto video frames, as shown in Fig.~\ref{fig:overview}.  Instead of using the solution-by-projection of \cite{li2020set}, we specify an Anchor-Constrained Viterbi (ACV) algorithm that efficiently approximates the MAP  ``all-color shortest path". Efficiency comes from significantly reducing the number of valid action segmentations by considering only those that include salient action parts, called anchor segments or anchors. Importantly, the ACV enables an end-to-end training,  as we use the generated pseudo-ground truth for computing the cross-entropy loss and regularization in the subsequent fully-supervised training of our HMM and MLP.

Given a training video, our approach begins by estimating saliency scores of all actions from the ground-truth set at every frame using the MLP. 
%
%The saliency score of an action at a frame is computed as a sum of log-softmax scores produced by the MLP over a temporal window centered at that frame. 
%
Then, for each action, we select the most salient frame in the video to represent a center of that action's anchor segment. %
%
%The length of anchor segments is estimated as a mean length of actions, i.e., the average of ratios of video lengths to the number of ground-truth actions. %
%
%When two anchor segments of two distinct actions happen to overlap, we keep the more salient anchor and for the other action select its second best salient frame and anchor. 
%
Finally, we approximate a globally optimal action segmentation by running the standard Viterbi algorithm over the densely connected video graph whose nodes are frames and colored edges represent action segments (see Fig.~\ref{fig:all-color}). Efficiency is achieved by considering only those paths in the video graph that pass through the anchor edges (i.e., segments). 

The resulting pseudo-ground truth is used in the second stage of our training, where we jointly learn parameters of the HMM and MLP using the cross entropy loss on the previously generated frame-wise pseudo-ground truth. We regularize this learning with a diversity loss aimed at maximizing a distance between saliency scores of distinct actions detected along the video. This is motivated by the following reasoning. For distinct actions, we expect that temporal patterns of their respective saliency scores across the frames are different. Thus, our diversity loss ensures that the saliencies of every action are sufficiently discriminative in time so as to facilitate action segmentation.

In evaluation, we address action segmentation and action alignment on the Breakfast, MPII Cooking2, Hollywood Extended datasets. %
%
%As in \cite{richard2018action,li2020set}, action alignment has access to the true set of actions present in  a test video, and the task is to accurately label frames with these actions. Note that this is more challenging than the common action alignment in  \cite{richard2017weakly,richard2018neuralnetwork}, where a test video additionally provides the known temporal ordering of actions present. 
%
Our experiments demonstrate that we outperform the state of the art on both tasks. 

In the following, Sec.~\ref{sec:Related Work} reviews related literature,  Sec.~\ref{sec:Problem Setup} defines our problem and models, Sec.~\ref{sec:Methods} specifies our ACV and regularization, Sec.~\ref{sec:inference} describes our inference on test videos, and Sec.~\ref{sec:Experiments} presents our results. 

%-------------------------------------------------------------------------
%-------------------------------------------------------------------------
\section{Related Work}\label{sec:Related Work}
This section puts our work in the context of recent approaches to weakly supervised action segmentation.  Fully supervised action segmentation requires frame-wise annotations (\eg, \cite{eyjolfsdottir2014detecting,kuehne2016end,lea2017temporal,lea2016segmental,lei2018temporal,ni2014multiple,rohrbach2012database,singh2016multi,tang2012learning,vo2014stochastic,yeung2016end,zhao2017temporal}), and hence is beyond our scope. Unsupervised action segmentation   is also beyond our scope (\eg \cite{kukleva2019unsupervised,sener2018unsupervised}), since our inference on test videos requires annotations of action sets present in training videos.

\textbf{Transcript-supervised Learning.} The temporal ordering of actions present in training videos is assumed known in \cite{bojanowski2014weakly,chang2019d3tw,chang2019d,ding2018weakly,huang2016connectionist,koller2016deep,koller2017re,kuehne2017weakly,Li_2019_ICCV,richard2017weakly,richard2018neuralnetwork}. For example,  Huang \etal \cite{huang2016connectionist} extended the connectionist temporal classification by taking into account a temporal coherence of features in consecutive frames. Bojanowski \etal \cite{bojanowski2014weakly} relaxed discriminative clustering based on a conditional gradient (Frank-Wolfe) algorithm for action alignment. Other approaches \cite{Li_2019_ICCV,richard2017weakly,richard2018neuralnetwork} formulated action segmentation with a statistical language model. Our training setting is less constrained, and hence more challenging. This motivates our simpler model choice in comparison with more complex deep architectures of prior work (\eg, recurrent network in \cite{Li_2019_ICCV,richard2018neuralnetwork}). 

\textbf{Set-supervised Learning.} The set of actions present in training videos is assumed known in
 \cite{jhuang2011large,liu2019completeness,liu2019weakly,nguyen2019weakly,paul2018w,shou2018autoloc,singh2017hide,soomro2012ucf101,sun2015temporal,wang2017untrimmednets,xu2019segregated,zhang2020adapnet,yu2019temporal,zhai2019action,Fayyaz_2020_CVPR}. For example, Shou \etal \cite{shou2018autoloc}  specified the outer-inner-contrastive loss for learning an action boundary detector,  Nguyen \etal  \cite{nguyen2019weakly} defined a background-aware loss to distinguish actions from the background, and Paul \etal \cite{paul2018w} proposed an action affinity loss for multi-instance learning. As shown in \cite{li2020set,richard2018action}, all of these approaches are specifically designed to address videos with sparse and rather few action occurrences. In contrast, we consider videos with dense actions and significantly more action occurrences. Similar to our approach, in \cite{Fayyaz_2020_CVPR}, temporal segments in the video are adaptively identified and labeled, but they do not use an HMM and do not sample valid sequences of actions for inference in testing.

In \cite{li2020set}, an HMM and a Viterbi-like algorithm are also used for estimating the frame-wise pseudo-ground truth, as in our approach. Their Set-Constrained Viterbi (SCV) algorithm uses a greedy post-processing step for ensuring that all actions from the set-level ground truth are included in the frame-wise pseudo-ground truth. This post-processing is heuristic, and thus suboptimal. In contrast, we formulate the ACV algorithm that efficiently approximates the ``all-color shortest path" and does not require post-processing. The ACV first estimates salient anchor frames and constructs an anchor-constrained graph, and then runs the vanilla Viterbi algorithm for optimization on the anchor-constrained graph. Unlike SCV, our ACV integrates both the estimation of salient anchors and anchor-constrained Viterbi in a unified end-to-end training.

%-------------------------------------------------------------------------
%-------------------------------------------------------------------------
\section{Our Problem Statement and Models}\label{sec:Problem Setup}
\subsection{Problem Formulation.}
A training video with length $T$ is represented by a sequence of frame features, $\bm{x} = [x_1,...,x_t,...,x_T]$, extracted in an unsupervised manner as in \cite{li2020set,richard2018action}. Each training video is annotated with a ground-truth set of actions  $C=\{c_1,\cdots, c_M\}\subseteq\mathcal{C}$, where $\mathcal{C}$ is the set of all actions, $\mathcal{C}=\{1,2,...,\mathcal{|C|}\}$. %$T$ and $M$ may vary across the training set, and there may be multiple instances of the same action in a training video.
Our goal is to find an optimal action segmentation of every training video, $(\hat{\bm{c}}, \hat{\bm{l}})$, where $\hat{\bm{c}}=[\hat{c}_1,...,\hat{c}_n,...,\hat{c}_{\hat{N}}]$ is the predicted temporal sequence of actions, $\hat{c}_n\in\mathcal{C}$, and $\hat{\bm{l}} = [\hat{l}_1,\cdots,\hat{l}_{\hat{N}}]$ are their corresponding temporal lengths.   $(\hat{\bm{c}}, \hat{\bm{l}})$ serves as our pseudo-ground truth for the subsequent pseudo-supervised training. 

\subsection{HMM}\label{sec:HMM}
We use an HMM to estimate the MAP $(\hat{\bm{c}}, \hat{\bm{l}})$  as
\begin{equation}
\arraycolsep=1pt
\begin{array}{lcl}
\label{eq:Markov}
(\hat{\bm{c}}, \hat{\bm{l}}) &=&  \displaystyle\argmax_{N, \bm{c},\bm{l}}\; p(\bm{c},\bm{l}|\bm{x})
%, \\
%& =&  
=\displaystyle\argmax_{N,\bm{c},\bm{l}}\; p(\bm{c})p(\bm{l}|\bm{c})p(\bm{x}|\bm{c},\bm{l}) \\
&= &  \displaystyle\argmax_{N,\bm{c},\bm{l}}\Big[p(c_1)\prod_{n=1}^{N-1} p({c_{n+1}}|c_n)\Big] \cdot \Big[\prod_{n=1}^{N}p(l_n|c_n)\Big] \\
& & \cdot \Big[\prod_{t=1}^{T}p(x_t|c_{n(t)})\Big],
 \end{array}
\end{equation}
where $p(c_1)$ is assumed equal for all action classes.

% In order to constrain and speed up our labeling search, we cut video into oversegmentation units of fixed $n_f$ frames. For all frames in a unit, labels are assumed to be same. Then the MAP $(\hat{\bm{c}}, \hat{\bm{l}})$ can be equivalently formulated as:
% \begin{align}\label{eq:unit Markov}
%     (\hat{\bm{c}}, \hat{\bm{l}}) = \argmax_{N, \bm{c},\bm{l}}\sum_{n=1}^{N-1}\log p({c_{n+1}}|c_n) + \sum_{n=1}^{N}\log p(l_n|c_n) + \sum_{u=1}^{N_u}e(u|c_u)
% \end{align}
% where $N_u = [\frac{T}{n_f}]$ is the number units, $e(u|c_u)$ is unit score of $u$th oversegmentation for class $c_u$. Specially, $e(u|c_u)$ is the sum of classwise log softmax score over all frames in the unit:
% \begin{align}\label{eq:unit score}
%     e(u|c_u) = \sum_{t=u \times n_f}^{(u+1) \times n_f - 1} \log p(x_t|c_u), 
% \end{align}

We study two versions of the HMM model defined in (\ref{eq:Markov}). One is the initial HMM, $\mathcal{H}=\mathcal{H}^0$, computed directly from the ground-truth action set, and the other is a refined HMM, $\mathcal{H}^i$, iteratively updated after $i$  training iterations starting from the initial $\mathcal{H}^0$. In the following, our notation for $\mathcal{H}^0$ and $\mathcal{H}^i$ uses 0 and $i$ in the superscript to indicate the training iterations, respectively.

\textbf{The Initial HMM.} The initial transition probability in \eqref{eq:Markov} is defined as
\begin{align}\label{eq:transition static}
    p^0({c_{n+1}}|c_n) = \frac{\#({c_{n+1}}, c_n)}{\#(c_n)},
    %\frac{\#({c_{n+1}}, c_n)}{\#(c_n)},
\end{align}
where $\#(\cdot)$ is the number of actions or pairs of actions in the ground truth. 

The initial action length in \eqref{eq:Markov} is modeled as the Poisson distribution:
\begin{align}
    p^0(l|c) = \frac{(\lambda_{c}^0)^{l}}{l!}e^{-\lambda_{c}^0},
    \label{eq:Poisson}
\end{align}
where $\lambda_{c}^0$ is the  expected temporal length of action $c \in \mathcal{C}$. For every $c$,  we estimate $\lambda_{c}^0$ such that the total length of all actions from $C$ is close to the video's length $T$. More formally, we estimate $\lambda_{c}^0$ by minimizing the following quadratic objective:
\begin{align}\label{eq:length static}
  \text{minimize} \quad  \sum_{v}\left(T_v - \sum_{c\in C_v} \lambda_c^0\right)^2, \nonumber \\
  \quad \text{s.t.}\quad  \text{for every } c: \; \lambda_c^0 > l_{\text{min}},
\end{align}
where $v$ is the index of videos, $l_{\text{min}}$  is the minimum action length ($l_{\text{min}} = 50$ frames in our experiments).

The initial likelihood at frame $t$ in  \eqref{eq:Markov} is estimated as:
\begin{align}
    p^0(x_t|c) \propto \frac{p(c|x_t)}{p^0(c)},\quad\quad p^0(c) = \frac{\sum_{v}T_v\cdot 1(c \in C_v)}{ \sum_{v}{T_{v}}}.
    \label{eq:likelihood_static}
\end{align}
where $p(c|x_t)$ is a softmax score of the MLP, and  $1(\cdot)$ is the indicator function. $p^0(c)$ in \eqref{eq:likelihood_static} is a percentage of the video footage having $c$ in the ground truth.

\textbf{The Refined HMM} is initialized to $\mathcal{H}^0$ as in \eqref{eq:transition static}--\eqref{eq:likelihood_static},
and then for each training video updated in $i$th training iteration based on the MAP assignment $(\hat{\bm{c}}^{i}, \hat{\bm{l}}^{i})$ for that video as
\begin{align}\label{eq:HMM dynamic iteration}
   & p^{i}(c_{n+1}|c_n) = p^{{i-1}}(c_{n+1}|c_n) \nonumber\\
      & \qquad + \frac{1}{V}\left(\frac{\#(\hat{c}_{n}^{i}, \hat{c}_{n+1}^{i})}{\#(\hat{c}_n^{i})} - p^{{i-1}}(c_{n+1}|c_n)\right), \\
    & \lambda_c^{{i}} =\lambda_c^{{i-1}} + \frac{1}{V}\left(\frac{\sum_{n=1}^{\hat{N}} \hat{l}_{n}^{i}\cdot 1(c=\hat{c}_{n}^{i})}{\sum_{n=1}^{\hat{N}} 1(c=\hat{c}_{n}^{i})} - \lambda_c^{{i-1}}\right),\\
    & p^{i}(c)= p^{{i-1}}(c) + \frac{1}{V}\left(\frac{\sum_{n=1}^{\hat{N}} \hat{l}_{n}^{i}\cdot 1(c=\hat{c}_{n}^{i})}{T} - p^{{i-1}}(c)\right).
    \label{eq:HMM dynamic iteration1}
\end{align}
where $V$ is the total number of training videos.  We use $\frac{1}{V}$ as our learning rate, since after we compute \eqref{eq:HMM dynamic iteration}--\eqref{eq:HMM dynamic iteration1} for all $V$  training videos, the updates become properly normalized. In  \eqref{eq:HMM dynamic iteration}, $\#(\hat{c}_n^{i})$ is the number of all segments labeled with $\hat{c}_n^{i}\in\mathcal{C}$, and $\#(\hat{c}_{n}^{i}, \hat{c}_{n+1}^{i})$ is the total number of pairs of two consecutive video segments labeled with $\hat{c}_{n}^{i}\in\mathcal{C}$ and $\hat{c}_{n+1}^{i}\in\mathcal{C}$, respectively. The update is done for each training video. This means that our choice of the update rate as $\frac{1}{V}$ for the updates in \eqref{eq:HMM dynamic iteration}--\eqref{eq:HMM dynamic iteration1} amounts to averaging over the pseudo-ground truth for all previously trained videos. In our experiments, the small update rate $\frac{1}{V}$  gives the ``stable'' and reliable updates in training.

%-------------------------------------------------------------------------
%-------------------------------------------------------------------------
\subsection{The MLP Network}\label{sec:MLP}
As in \cite{richard2018action}, our two-layer MLP  scores frames for all action classes. The MLP's hidden layer $\bm{h} \in \mathbb{R}^{n_h}$, $n_h=256$, is input to the binary classifier ${f}_c(\bm{x})$, for every $c\in \mathcal{C}$, as
\begin{align}
    \bm{h}(\bm{x}) &= \text{ReLU}(\bm{W}^1\bm{x} + \bm{b}^1),
    \label{eq:hidden}\\
    {f}_c(\bm{x}) &= \sigma(\bm{W}^{\top}_c\bm{h}(\bm{x}) + b_c), \; c\in \mathcal{C},
    \label{eq:secondlayer}
\end{align}
where $\bm{W}_c \in \mathbb{R}^{n_h}$, ${b}_c \in \mathbb{R}^{1}$, and $\sigma$ is the sigmoid function. 

$p(c|x)$ in \eqref{eq:likelihood_static} is estimated as a softmax score of the MLP's binary classifier $f_c(x)$ at frame $x$. To ensure reliable $p(c|x)$, we pretrain the MLP as in \cite{richard2018action}. Specifically, we pretrain each $f_c(x)$, $c\in \mathcal{C}$, using  multi-instance learning.

%%%%%%%%%%%%%%%%%%%%%%%%%%%%%%%%%%%%%%%%%%%%%%%%%%%%%%%%%%%%%%

\section{Our Set-Supervised Training}\label{sec:Methods}
Our training consists of two stages. First, we generate the MAP pseudo-ground truth $(\hat{\bm{c}}, \hat{\bm{l}})$ for every training video with our ACV algorithm. 
Second, the pseudo-ground truth is used to estimate the cross-entropy loss and regularization for updating the HMM and MLP. Below, we first formulate our ACV, and then the loss and regularization.   

\begin{figure*}[ht]
\begin{center}
\includegraphics[width=0.66\linewidth]{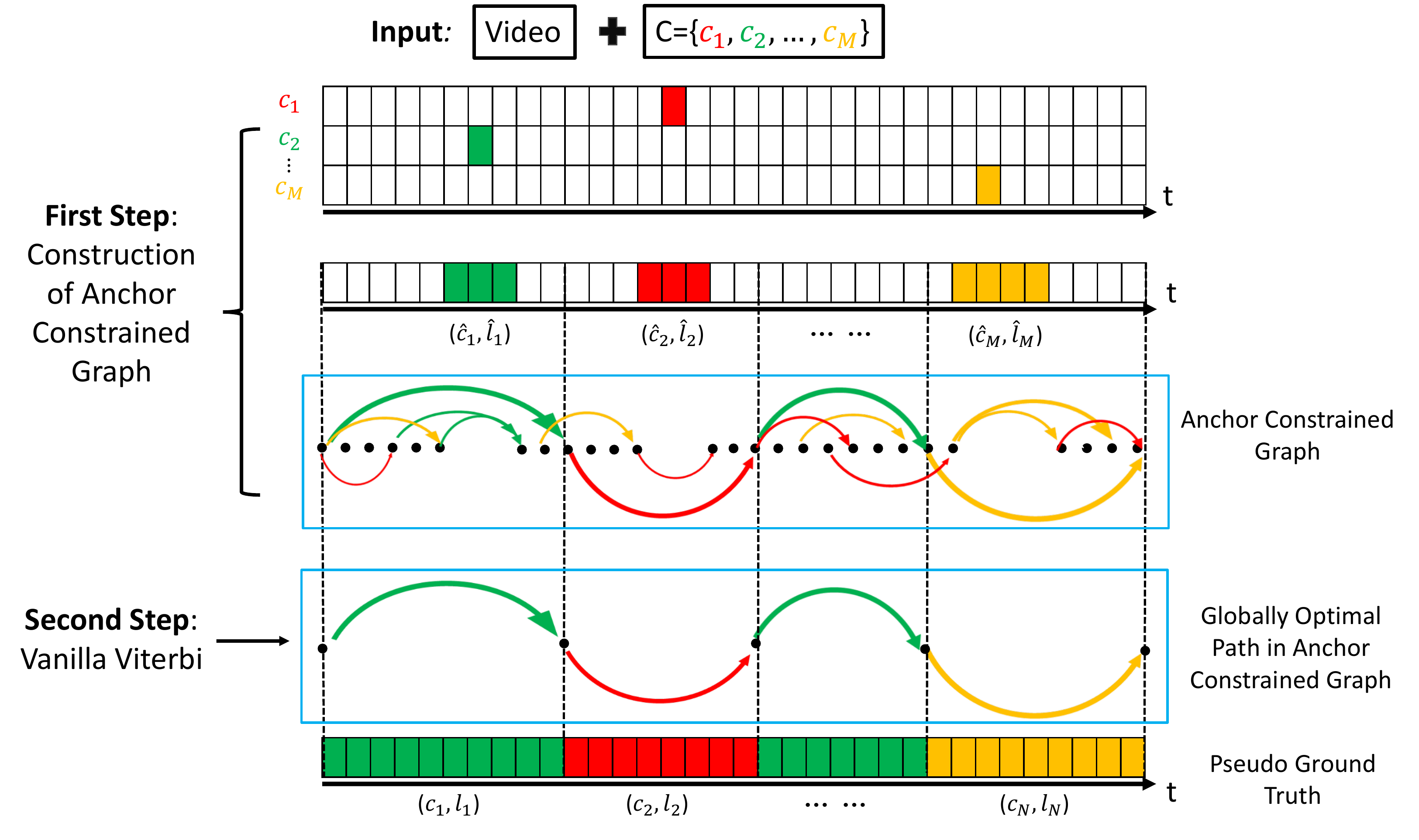}
\end{center}
  \caption{Our ACV consists of two steps. First, we compute saliency scores of videos frames for all actions $c$ from the ground truth $C$, and for each $c\in C$  select the most salient frame, called anchor. Each anchor of action $c$ is taken as a center of the corresponding anchor segment whose length is proportional to the expected length $\lambda_c$ of $c$. The anchor segments are used to construct an anchor-constrained graph  which efficiently represents all valid action segmentations by directed paths in the graph. Second, we run the Viterbi algorithm to efficiently identify the MAP path in the anchor-constrained graph.}
\label{fig:ACV}
\end{figure*}

\subsection{Anchor-Constrained Viterbi}
Given a training video and its ground truth $C$, our goal is to predict the MAP $(\hat{\bm{c}}, \hat{\bm{l}})$, such that every action $c\in C$ occurs at least once in $\hat{\bm{c}}$. As shown in \cite{DBLP:journals/corr/BilgeCGSAE15}, this  ``all-color shortest path" problem is NP-hard. Our ACV algorithm is an efficient approximation that consists of two related steps, as illustrated in Fig.~\ref{fig:ACV}, where the first step constructs an anchor-constrained graph aimed at representing all valid action segmentations, and the second step finds the MAP valid segmentation in the graph.

{\bf The first step} computes saliency  scores $\bm{S} \in \mathbb{R}^{M \times T}$ of all actions  $c$ in the ground truth $C$, $M=|C|$, at every frame $t=1,\dots,T$ using a sliding window of length $(2\tau+1)$ as
\begin{equation}\label{eq:salient score}
    \bm{S}[c,t] = \sum_{u=-\tau}^{\tau} [\log f_c(x_{t+u}) - \min_{c' \in C} \log f_{c'}(x_{t+u})],
\end{equation}
where $f_c(x_{t})$ is given by \eqref{eq:secondlayer}, and $\tau=15$ frames in our experiments. By subtracting $\min_{c'} \log f(c')$ in \eqref{eq:salient score}, we specify frame saliency as a difference between the highest and lowest softmax scores, instead of considering only the highest softmax scores. We have empirically observed that the latter leads to a high false-positive rate, as some frames have significantly higher softmax scores than others.

For every $c \in C$, we select its most salient frame $t_c$ and take it to represent a center of $c$'s anchor segment.  This gives $M=|C|$ anchor segments in the video. The anchors are later used to constrain the search for an optimal ``all-color shortest path" by considering only candidate action segmentations that include the anchors. A time interval centered at every $t_c$, $[t_c-\frac{\alpha}{2}\lambda_c, t_c+\frac{\alpha}{2}\lambda_c]$ is estimated as $c$'s anchor segment, where $\lambda_c$ is the expected length of action $c$ given by \eqref{eq:length static}, and $\alpha\in (0,1]$ is a constant. We empirically found that $\alpha=0.6$ is optimal. Since $\alpha\in (0,1]$, the  anchor segments are set to have smaller lengths than the expected lengths of the corresponding actions in order to allow some flexibility of action boundary learning.
For every two anchor segments that happen to overlap, we keep the more salient anchor, and for the other action we iteratively select a new anchor as that action's second best or less salient frame until there is no overlap.

The anchor segments are used to construct a densely connected video segmentation graph, as illustrated in Fig.~\ref{fig:ACV}. In the graph, directed edges represent candidate actions along the video such that each edge is constrained to include one and only one of the anchor segments. Nodes in the graph represent frames that are candidate boundaries (or cuts) between actions, such that no cut can occur within the anchor segments.  A directed path in the graph represents a candidate action segmentation. Therefore, by construction, every path in the anchor-constrained graph includes all actions from the ground truth $C$ at least once, and thus represents a valid candidate for approximating our   ``all-color shortest path" problem.

%In some extreme cases, the interval between two salient frames $\tilde{t}_i$ and $\tilde{t}_{i+1}$ are not large enough to hold two half action anchors, i.e. $(\tilde{t}_{i+1} - \tilde{t}_{i}) < \alpha(\lambda_{\tilde{c}_i} + \lambda_{\tilde{c}_{i+1}})/2$, then we assign boundary between $\tilde{t}_i$ to $\tilde{t}_{i+1}$ according to their mean action lengths:
%\begin{align}
%   \tilde{e}_i = \tilde{t}_i + [\frac{\lambda_{\tilde{c}_i}}{\lambda_{\tilde{c}_i}+\lambda_{\tilde{c}_{i+1}}}(\tilde{t}_{i+1} - \tilde{t}_{i})],
%  \quad \tilde{s}_{i+1} = \tilde{e}_i + 1
%\end{align}

% \begin{equation}
% c_{u}=\left\{
% \begin{aligned}
% \tilde{c}_i, \quad &\text{if } \ \tilde{c}_i \le u \le \tilde{c}_i + \frac{\lambda_{\tilde{c}_i}}{\lambda_{\tilde{c}_i}+\lambda_{\tilde{c}_{i+1}}}(\tilde{u}_{i+1} - \tilde{u}_{i})\\
% \tilde{c}_{i+1}, \quad &\text{if } \ \tilde{c}_i + \frac{\lambda_{\tilde{c}_{i+1}}}{\lambda_{\tilde{c}_i}+\lambda_{\tilde{c}_{i+1}}}(\tilde{u}_{i+1} - \tilde{u}_{i}) < u \le \tilde{c}_{i+1}
% \end{aligned}
% \right.
% \end{equation}

%Therefore, we obtain the action anchor for all actions
%\begin{align}\label{eq:action anchor}
%     [(\tilde{c}_1,\tilde{l}_1), \cdots, (\tilde{c}_{N+1},\tilde{l}_{N+1})] = [(\tilde{c}_1,\tilde{s}_1,\tilde{e}_1), \cdots, (\tilde{c}_{N+1},\tilde{s}_{N+1},\tilde{e}_{N+1})]
%\end{align}
%where $\tilde{l}_i, \tilde{s}_i,\tilde{e}_i$ are length, start frame and end frame for the $i^{th}$ action anchor. 

{\bf The second step} can be viewed as efficiently finding an optimal path in the anchor-constrained graph (see Fig.~\ref{fig:ACV}). For this, we use the Viterbi algorithm that predicts the MAP $(\hat{\bm{c}}, \hat{\bm{l}})$ in this graph. From \eqref{eq:Markov}--\eqref{eq:HMM dynamic iteration1} and \eqref{eq:secondlayer}, the Viterbi is formalized as
\begin{align}\label{eq:ThirdStep}
    (\hat{\bm{c}}, \hat{\bm{l}}) = \argmax_{\substack{N,\bm{c},\bm{l}\\ \bm{c}\in C^N}}&\sum_{n=1}^{N-1}\log p({c_{n+1}}|c_n) + \sum_{n=1}^{N}\log p(l_n|c_n) \nonumber \\
   & + \sum_{t=1}^{T} \log p(x_t|c_{n(t)}).
\end{align}
under the constraint that all segments $l_n$ in \eqref{eq:ThirdStep} must be selected from the edges of the anchor-constrained graph.

% \begin{align}
%      (\hat{\bm{c}}, \hat{\bm{l}}) &= \argmax_{\substack{M_i,\bm{c}_i,\bm{l}_i\\ \bm{c}_i\in C}}\; p(\bm{c}_i,\bm{l}_i|\bm{f}[\tilde{e}_{i}+1:\tilde{s}_{i+1}-1]), \nonumber \\
%           & =  \argmax_{\substack{M_i,\bm{c}_i,\bm{l}_i\\ \bm{c}_i\in C}} \Big(\prod_{n=0}^{M_i}p(c_{n+1}|c_n)\Big)\cdot \Big(\prod_{n=1}^{M_i}p(l_n|c_n)\Big) \nonumber \\
%           &  \cdot \Big(\prod_{t=\tilde{e}_i + 1}^{\tilde{s}_{i+1} - 1}\frac{p(\bm{f}[c_{n(t)},t])}{p(c_{n(t)})}\Big),
%           \label{eq:ThirdStep}
% \end{align}
% where we set $c_0=\tilde{c_i}$, $c_{M_i+1}=\tilde{c_{i+1}}$ as the left and right action anchors, $p(\bm{f}[c_{n(t)},t])$ denotes the soft-max score for class $c_n\in C$ at frame $t$.

%-------------------------------------------------------------------------

The Viterbi begins by computing: (i) Frame likelihoods $p(x_t|c)$ for $t=1,\dots,T$, using the MLP softmax scores  as in \eqref{eq:likelihood_static} and \eqref{eq:secondlayer}, and (ii) Transition probabilities $p(c|c')$ for all $c, c'\in\mathcal{C}$ as in \eqref{eq:transition static}. 
Then, the Viterbi uses a recursion to efficiently compute $(\hat{\bm{c}_i}, \hat{\bm{l}_i})$ in \eqref{eq:ThirdStep}. Let $v(c,t)$ denote the maximum score for all action sequences $\bm{c}_t$  ending with action $c \in C$ at video frame $t$, where a total sum of the mean action lengths along each sequence until $t$ is less than $1.5 T$, $(\sum_{c'\in\bm{c}_t} \lambda_{c'})<1.5 T$. From \eqref{eq:ThirdStep},  $v(c,t)$ can be recursively estimated as
\begin{align}
    v(c,t)  = \max_{\substack{t'<t\\  c'\neq c \\ c'\in C}}\; &\Big[ v(c',t')+ \log p(t{-}t'|c) + \log p(c|c') \nonumber \\
     & + \sum_{k = t'+1}^{t} \log p(x_k|c)\Big],
\end{align}
where $p(t{-}t'|c)$ is the action length likelihood given by \eqref{eq:Poisson}, $p(c|c')$ is the transition probability from $c'$ to $c$ as in \eqref{eq:transition static}, $p(x_k|c)$ is the likelihood at frames $k=(t'+1),\dots, t$ given by \eqref{eq:likelihood_static}. During the recursion, we enforce that time intervals $(t-t')$ must respect edges in the anchor-constrained graph (i.e., must include anchor segments). Finally, the optimal path can be back-tracked from  $\max_{c\in C}\; v(c,T)$, resulting in an optimal action segmentation $(\hat{\bm{c}}, \hat{\bm{l}})$. 

In our implementation, we improve efficiency by discarding early unrealistic candidate solutions with too many action changes along the path -- i.e., every subpath ${\bm{c}_t}$ whose accumulated mean length exceeds $1.5$ times the current temporal length, $\sum_{c\in {\bm{c}_i}}\lambda_c> 1.5 t$.

The predicted action segmentation is used as a pseudo-ground truth to estimate the cross-entropy loss for pseudo fully-supervised training of the HMM and MLP, as explained in the following section.

% In training, we discard the predicted action sequence $\tilde{\bm{c}}$  whose accumulated mean length exceeds the video length $T$, $\sum_{c\in\tilde{\bm{c}}}\lambda_c>T$, and select the next best $\tilde{\bm{c}}$ in  $\bm{s}$. 

% By concatenating pseudo-labels in each interval and all action anchor segments, we get final pseudo-labels for the entire video as $\hat{\bm{c}} = [\tilde{c}_0, \hat{\bm{c}_0}, \cdots, \tilde{c}_N, \hat{\bm{c}_N}, \tilde{c}_{N+1}]$ and $\hat{\bm{l}} = [\tilde{l}_0, \hat{\bm{l}_0}, \cdots, \tilde{l}_N, \hat{\bm{l}_N}, \tilde{l}_{N+1}]$.

%-------------------------------------------------------------------------
\subsection{The Cross-Entropy Loss}
The ACV inference result, $(\hat{\bm{c}}, \hat{\bm{l}})$, incurs the following binary cross-entropy loss over all action classes as
\begin{align}\label{eq:cross entropy}
    \mathcal{L}_{CE} = - \frac{1}{T}\big[\sum_{t=1}^{T}\sum_{c\in \mathcal{C}}1(\hat{c}_t=c) \log p(c|x_t) \nonumber \\
    + 1(\hat{c}_t\neq c) \log (1 - p(c|x_t))\big], \;\hat{c}_{t}\in \hat{\bm{c}}
\end{align}
where $p(c|x_t)$ is the MLP's softmax score for class $c$ at frame $t$, given by \eqref{eq:secondlayer}.

% After the ACV infers the pseudo-ground truth of a given training video,  $(\hat{\bm{c}}, \hat{\bm{l}})$, we compute the incurred binary cross-entropy loss over all action classes as
% \begin{align}\label{eq:cross entropy}
%     \mathcal{L}_{CE} = - \sum_{i=1}^{\mathcal{C}} \sum_{t=1}^{T}\log p(\hat{c}_{n(t)}|x_t), \;\hat{c}_{n(t)}\in \hat{\bm{c}}
% \end{align}
% where $p(\hat{c}_{n(t)}|x_t) = p(\bm{f}[\hat{c}_{n(t)},t])$ is the soft-max score of the two-layer neural network for the $n$th predicted class $\hat{c}_{n(t)}\in \hat{\bm{c}}$ at frame $t$.

%-------------------------------------------------------------------------

% \begin{figure}[t]
% \begin{center}
% \includegraphics[width=0.95\linewidth]{n-pair.pdf}
% \end{center}
%   \caption{A simple example of estimating three cosine distances that are used for computing the n-pair loss. Given video $v$ with only two classes $\{c,r\}$ and another video $v'$ with also two classes $\{c,s\}$, we first compute their average class features: $\{\bar{h}_v^c,\bar{h}_v^r\}$ and $\{\bar{h}_{v'}^c, \bar{h}_{v'}^s\}$. Then, we estimate the cosine distance $d_{vv'}^{cc}$ of $\bar{h}_v^c$ and $\bar{h}_{v'}^c$ for the shared class $c$, and the cosine distances $d_{vv'}^{rc}$ and $d_{vv'}^{cs}$ of the video features for the non-shared classes $r$ and $s$.}
% \label{fig:n-pair}
% \end{figure}

\subsection{Diversity Loss}
We regularize our learning by using the action saliency scores estimated for all actions $c$ at every frame, $\bm{S}[c,1{:}T] = [\bm{S}[c,1],..., \bm{S}[c,t],...  \bm{S}[c,T]]$, where $\bm{S}[c,t]$ is  given by \eqref{eq:salient score}. Distinct actions are expected to exhibit different saliency patterns along the video in $\bm{S}[c,1{:}T]$. Therefore, we specify a diversity loss, $\mathcal{L}_{DIV}$, for regularizing our learning so as to maximize a distance between every pair of $\bm{S}[c,1{:}T]$ and $\bm{S}[c',1{:}T]$, $c\ne c'$.

For a training video with the ground truth $C$, we define $\mathcal{L}_{DIV}$ over all pairs of distinct actions $(c,c')\in C\times C$, $c\ne c'$, as an average normalized correlation between $\bm{S}[c,1{:}T]$ and $\bm{S}[c',1{:}T]$  as
\begin{align}\label{eq:DIV}
    \mathcal{L}_{DIV} = \frac{1}{M(M-1)}\sum_{\substack{(c,c')\in C{\times} C \\ c \neq c'}}\frac{\bm{S}[c,1{:}T]^{\top}\bm{S}[c',1{:}T]}{\|\bm{S}[c,1{:}T]\|_2\|\bm{S}[c',1{:}T]\|_2}.
\end{align}
The normalization in \eqref{eq:DIV} is necessary, since the number of actions varies across training videos. 

% SINISA: this sentence is not precise, and does not bring any information.
%Without this normalization, the diversity loss would be unstable in practice.

Our total loss is a sum of $\mathcal{L}_{CE}$ and $\mathcal{L}_{DIV}$:
\begin{align}
    \mathcal{L} = \mathcal{L}_{CE} + \beta\mathcal{L}_{DIV},
\end{align}
where we experimentally found that $\beta=0.4$ is optimal.

%-------------------------------------------------------------------------
\subsection{Complexity of Our Training}
Complexity of ACV is $O(T^2|\mathcal{C}|^2)$, where $T$ is the length of a training video, and $|\mathcal{C}|$ is the number of all action classes. This mainly comes from the Viterbi in  ACV, whose complexity is $O(T^2|\mathcal{C}|^2)$. Complexity of the first step of ACV is $O(T|\mathcal{C}|) < O(T^2|\mathcal{C}|^2)$. Complexity of the diversity loss is $O(T|\mathcal{C}|^2)$. Therefore, the total complexity of our training is $O(T^2|\mathcal{C}|^2)$. The same training complexity is reported in \cite{li2020set}.

%-------------------------------------------------------------------------
\section{Inference on a Test Video}\label{sec:inference}
In inference on a test video, we use a Monte Carlo sampling, as in \cite{li2020set,richard2018action}. Given a test video of length $T$, we first randomly select one action set $C$ from the ground truths seen in training. Then, we sequentially sample from $C$ a temporal sequence of actions $\bm{c}$ until  $\sum_{c\in\bm{c}}\lambda_c > T$. We repeat this sequential sampling when $\bm{c}$ is discarded for not including all actions from $C$. In this way, we end up generating K=1000 valid sequences $\mathbb{C}=\{\bm{c}\}$. For every valid $\bm{c}\in \mathbb{C}$, we use the standard dynamic programming to infer optimal action segmentation in the test video  by maximizing the HMM's posterior, $(\bm{c},\hat{\bm{l}})=\argmax_{\bm{l}} p(\bm{c},\bm{l}|\bm{x})$. Among the K sequences in $\mathbb{C}$, we select the action segmentation with the maximum posterior as our final solution, $(\bm{c}^*,\bm{l}^*)=\argmax_{\bm{c}\in \mathbb{C}} p(\bm{c},\hat{\bm{l}}|\bm{x})$.

%-------------------------------------------------------------------------
  Complexity of our inference is $O(T^2|\mathcal{C}|K)$, where $T$ is the length of a test video, $|\mathcal{C}|$ is the number of action classes, $K$ is the number of generated action sequences. The same complexity is reported in \cite{li2020set}.

%-------------------------------------------------------------------------

\section{Experiments}\label{sec:Experiments}

{\bf Datasets.} As in \cite{li2020set,richard2018action}, for evaluation, we address the tasks of action segmentation and action alignment on three datasets, including Breakfast \cite{kuehne2014language},   Hollywood Extended (Ext)  \cite{bojanowski2014weakly}, and MPII Cooking 2  \cite{rohrbach2016recognizing}. 
{\em Breakfast} consists of 1,712 videos of breakfast cooking comprising 48 actions. There are on average 6.9 action instances in each video. As in \cite{kuehne2014language}, a mean of frame accuracy (Mof) over the same 4-fold cross validation is reported.
{\em Hollywood Ext} consists of 937 video clips showing a total of 16 actions. There are on average 2.5 action instances in each video. As in \cite{kuehne2017weakly}, we perform the same 10-fold cross validation and report the mean intersection over detection (IoD) averaged over the 10 folds, defined as $\text{IoD} = |GT\cup D|/|D|$, where $GT$ and $D$ denote the ground truth and detected action segments with the largest overlap.
{\em MPII Cooking 2} consists of 273 videos showing cooking activities. There are 67 action classes. Following \cite{rohrbach2012database}, the same training and testing split is used. We report the midpoint hit metric, i.e. a ratio of the midpoint of a correctly detected segment in the ground-truth segment, as in \cite{rohrbach2012database}.

{\bf Features.} For a fair comparison, our approach is evaluated on the same unsupervised video features as in \cite{li2020set,richard2018action}. For all three datasets, we extract frame-wise Fisher vectors of improved dense trajectories \cite{wang2013action} extracted over a sliding window of 20 frames. Our features are 64-dimensional projections by PCA of the Fisher vectors.

{\bf Training.} Our training starts from pretraining binary classifiers in \eqref{eq:secondlayer} with a multi-instance learning, as in \cite{richard2018action}. Then we train our model for a total number of 100,000 iterations, where we randomly select one video in each training iteration. The learning rate is initialized to 0.01 and reduced to 0.001 at the 10,000th iteration. The refined HMM's parameters are updated after each iteration as in \eqref{eq:HMM dynamic iteration}--\eqref{eq:HMM dynamic iteration1}.

{\bf Ablations.} We consider the following variants of our approach for evaluating the effect of each component:
\begin{itemize} [itemsep=0pt,topsep=-5pt, partopsep=-1pt]
\item ACV = Our full approach with the diversity loss, updated transition probabilities, updated mean lengths, updated class priors given by \eqref{eq:HMM dynamic iteration}--\eqref{eq:HMM dynamic iteration1}.
\item ACVinitial = ACV with the initial transition probabilities, initial mean lengths, initial class priors given by \eqref{eq:transition static}, \eqref{eq:length static}, \eqref{eq:likelihood_static}. 
\item ACVnoreg = ACV without any regularization. 
\item ACV+Npair= ACV without the diversity regularization, but with the N-pair loss regularization.
\end{itemize}

\begin{table}[ht]
\begin{center}
\begin{tabular}{l c c c}
\hline
  & Breakfast & Cooking2 & Holl.Ext \\
 Model  & (\textit{Mof}) & (\textit{midpoint}) & (\textit{IoD}) \\
\hline
\text{(Set-supervised)} \\
Action Set \cite{richard2018action} & 23.3 & 10.6 & 9.3\\
SCT \cite{Fayyaz_2020_CVPR}our features & 26.6 & 14.3 & 17.7\\
SCV \cite{li2020set}  & 30.2 & 14.5 & 17.7\\
Our ACV& {\bf 33.4} & {\bf 15.5} & {\bf 20.9}\\
\hline
\multicolumn{2}{l}{\text{(Transcript-supervised)} }\\
OCDC \cite{bojanowski2014weakly} & 8.9 & - & - \\
HTK \cite{kuehne2017weakly} & 25.9 & 20.0 & 8.6\\
CTC \cite{huang2016connectionist} & 21.8 & - & -\\
ECTC \cite{huang2016connectionist} & 27.7 & - & -\\
HMM+RNN \cite{richard2017weakly} & 33.3 & - & 11.9\\
TCFPN \cite{ding2018weakly} & 38.4 & - & 18.3\\
NN-Viterbi \cite{richard2018neuralnetwork} & 43.0 & - & - \\
D3TW \cite{chang2019d3tw} & 45.7 & - & - \\
CDFL \cite{Li_2019_ICCV} & 50.2 & - & 25.8 \\
\hline
\end{tabular}
\end{center}
\caption{(Top) Our ACV outperforms the state-of-the-art set-level supervised approaches in terms of Mof, Midpoint, IoD. (Bottom) We compare ACV with weakly supervised approaches which have access to the true temporal ordering of actions in training. Our ACV achieves comparable results and outperforms some weakly supervised approaches. The dash means ``not reported''. When SCT \cite{Fayyaz_2020_CVPR} uses different 2048-dimensional frame features for Breakfast, their Mof of 30.4 is still inferior to ours.}
\label{Table:action segmentation}
% \vskip -0.1in
\end{table}

\subsection{Action Segmentation}
This section presents our action segmentation on the three datasets. As shown in Tab.~\ref{Table:action segmentation}, our ACV outperforms the state of the art by 3.2\% on Breakfast, 1.0\% on Cooking 2, and 3.2\% on Hollywood Ext respectively. In Tab.~\ref{Table:action segmentation}, we also compare ACV with prior work that has access to stronger supervision in training, where ground truth is a transcript that additionally specifies the temporal ordering of actions. ACV outperforms some recent transcript-supervised approaches.
%, despite weaker supervision in training. 

\begin{figure}[ht]
\begin{center}
\includegraphics[width=0.95\linewidth]{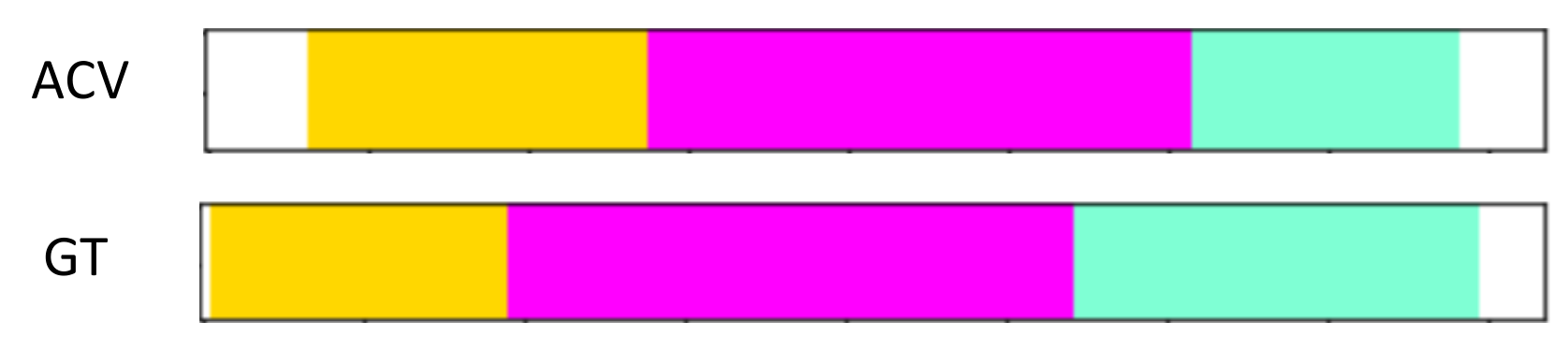}
\end{center}
\caption{Qualitative result of action segmentation on a sample test video \textit{P15\_cam01\_P15\_sandwich} from Breakfast. Top row: ACV result. Bottom row: ground truth with the color-coded action sequence \{\textcolor{mygold}{cut\_bun}, \textcolor{mymagenta}{smear\_butter}, \textcolor{myaquamarine}{put\_toppingOnTop}\}. We mark the background frames in white. In general, ACV can detect true actions present in videos, but may miss their true locations.}
\label{fig:seg_example}
\end{figure}

Fig.~\ref{fig:seg_example} qualitatively illustrates our action segmentation on a test video \textit{P15\_cam01\_P15\_sandwich} from Breakfast.  As can be seen, in general, ACV  can detect true actions present in videos, but may miss the true locations of their start and end frames.

\begin{table}[ht]
\begin{center}
\begin{tabular}{l c c c}
\hline
 & Breakfast & Cooking 2 & Holl.Ext \\
Model parameters & (\textit{Mof}) & (\textit{midpoint}) & (\textit{IoD}) \\
\hline
ACV+ground truth & 36.3 & 16.8 & 22.1 \\
\hline
ACVinitial & 31.9 & 15.0 & 19.1\\
ACV & 33.4 & 15.5 & 20.9\\
\hline
\end{tabular}
\end{center}
\caption{Evaluation of ACV when using the HMM with the ground-truth parameters, initially estimated parameters, and updated refined parameters.}
\label{Table:Comparison of length models.}
% \vskip -0.2in
\end{table}

{\bf Ablation --- Initial vs Refined.} Tab.~\ref{Table:Comparison of length models.} shows the results our ACV with different HMM parameters. Our ACV with updated HMM parameters outperforms the variant with initial HMM parameters -- ACVinitial. The table also shows the upper-bound performance of ACV (ACV+ground truth) when the mean action length $\lambda_c$, transition probability $p(c|c')$, and class priors $p(c)$ are estimated directly from ground-truth labels of video frames. As can be seen in Tab.~\ref{Table:Comparison of length models.}, our iterative updating of model parameters, given by  \eqref{eq:HMM dynamic iteration}, in ACV is reasonable, since ACV+ground truth outperforms ACV by only 2.9\% on Breakfast.

{\bf Ablation --- Regularization.} Tab.~\ref{Table:Regularizations Comparision.} compares our ACV with the other variants: ACVnoreg and ACV+Npair, where the latter uses the same loss as in \cite{li2020set}. As the N-pair loss needs two videos, as in \cite{li2020set}, we sample two videos that share common actions in each iteration. In contrast, our ACV follows a more general training strategy where only one randomly selected video is needed. The table shows that ACV gives the best performance.

\begin{table}[hb]
\centering
\begin{tabular}{l c c c}
\hline
 & Breakfast & Cooking 2 & Holl.Ext \\
Model  & (\textit{Mof}) & (\textit{midpoint hit}) & (\textit{IoD}) \\
\hline
ACVnoreg & 32.1 & 15.1 & 19.9\\
ACV+Npair & 32.4 & 15.1 & 20.2 \\
ACV & {\bf 33.4} & {\bf 15.5} & {\bf 20.9}\\
\hline
\end{tabular}
\vskip 0.1in
\caption{Evaluation of ACV w/out and w/ regularization.}
\label{Table:Regularizations Comparision.}
% \vskip -0.2in
\end{table}

\begin{figure}[ht]
\begin{center}
\includegraphics[width=0.95\linewidth]{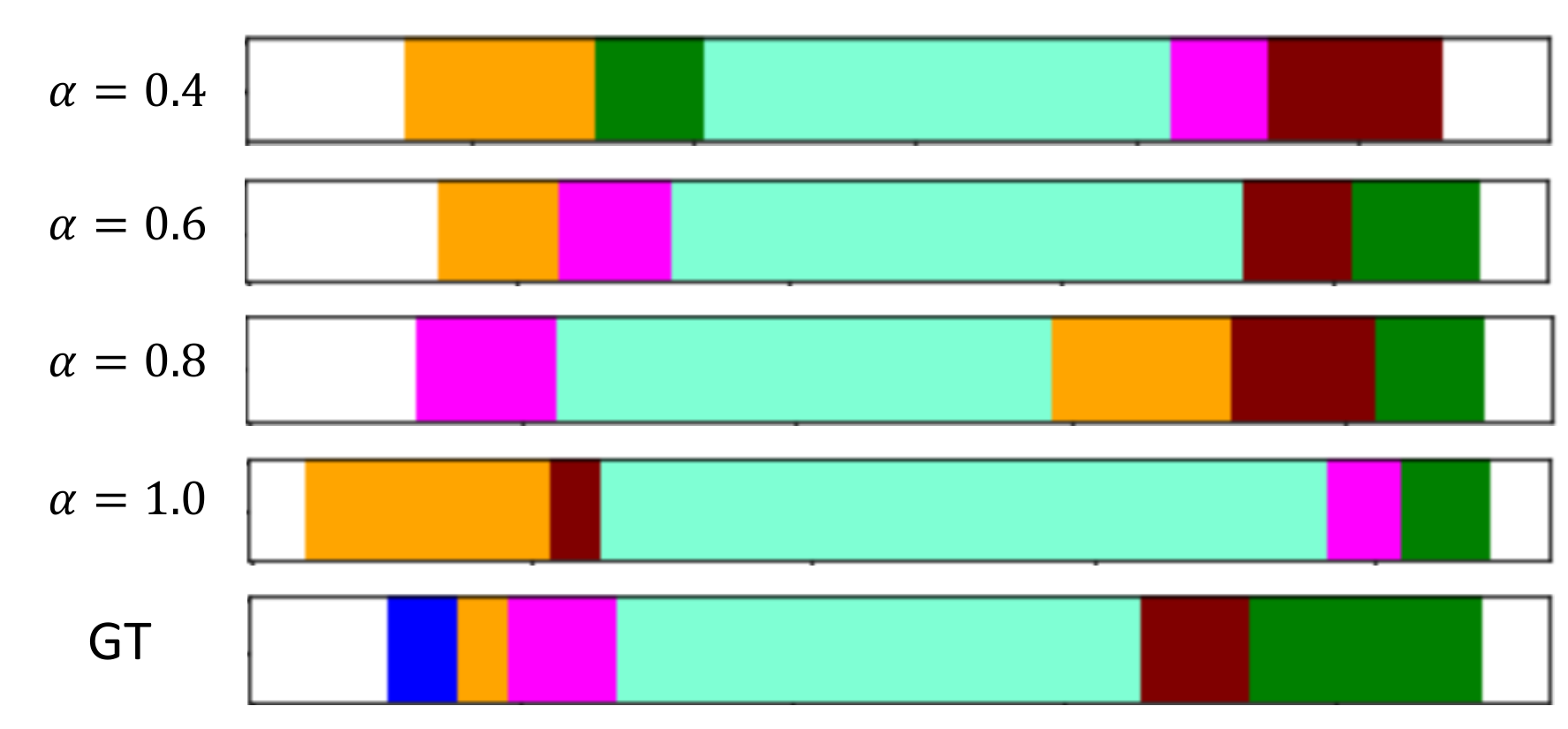}
\end{center}
  \caption{The effect of choosing different lengths of anchor segments controlled by $\alpha$ on the test video \textit{P07\_webcam01\_P07\_juice} from Breakfast. The predicted and ground-truth (GT) action sequences are color-coded \{\textcolor{myblue}{take\_plate},  \textcolor{myorange}{take\_knife},  \textcolor{mymagenta}{cut\_orange}, \textcolor{myaquamarine}{squeeze\_orange},  \textcolor{mymaroon}{take\_glass}, 
  \textcolor{mygreen}{pour\_juice}\}. The background frames are marked white. ACV with $\alpha=0.6$ gives the best performance.}
\label{fig:anchor_length_ablation}
\end{figure}

{\bf Ablation --- Anchor length.} Tab.~\ref{Table:Anchor Length Comparision.} shows the effect of input parameter $\alpha$ which controls the length of anchor segments. As can be seen, ACV with $\alpha=0.6$ performs the best. This supports our design choice to allow flexibility  in estimation of action lengths in the Viterbi  (as opposed when $\alpha=1$). Fig.~\ref{fig:anchor_length_ablation} visualizes our sensitivity to the choice of $\alpha$ on the test video \textit{P07\_webcam01\_P07\_juice} from Breakfast. As can be seen, ACV with $\alpha=0.6$ gives the best result. 

\begin{table}[hb]
\centering
\begin{tabular}{l c c c}
\hline
 & Breakfast & Cooking 2 & Holl.Ext \\
Model  & (\textit{Mof}) & (\textit{midpoint}) & (\textit{IoD}) \\
\hline
$\alpha=0.4$ & 32.5 & 15.2 & 20.1\\
$\alpha=0.6$ & {\bf 33.4} & {\bf 15.5} & {\bf 20.9}\\
$\alpha=0.8$ & 31.7 & 15.0 & 19.5 \\
$\alpha=1.0$ & 30.5 & 14.7 & 18.7 \\
\hline
\end{tabular}
\vskip 0.1in
\caption{Sensitivity of ACV to anchor-segment length.}
\label{Table:Anchor Length Comparision.}
\end{table}

Tab.~\ref{Table:Evaluation of salient anchors} shows accuracy of the first step of our ACV in terms of intersection over detection (IoD). That is we estimate IoD between the estimated anchor segments and ground truth action intervals. The table shows that $\alpha=0.6$ gives the highest IoD, and that accuracy improves as the number of training iterations grows.

\begin{table}[bt]
\begin{center}
\begin{tabular}{l c c c c c c}
\hline
Iteration & 0k & 20k & 40k & 60k &	80k & 100k \\
\hline
$\alpha=0.4$ & 19.1 & 27.5 & 30.0 &	35.5 &	37.0 &	37.8\\
$\alpha=0.6$ & 19.4 & 28.7 & 32.2 &	36.6 &	39.0 &	39.2\\
$\alpha=0.8$ & 18.7 & 27.3 & 29.9 &	35.0 &	36.8 &	37.1\\
$\alpha=1.0$ & 18.1 & 25.1 & 29.4 &	33.7 &	35.2 &	35.9\\
\hline
\end{tabular}
\end{center}
\caption{IoD evaluation of salient anchors during training on Breakfast dataset.}
\label{Table:Evaluation of salient anchors}
% \vskip -0.2in
\end{table}

\subsection{Action Alignment Given Action Sets}
\begin{figure}[ht]
\begin{center}
\includegraphics[width=0.95\linewidth]{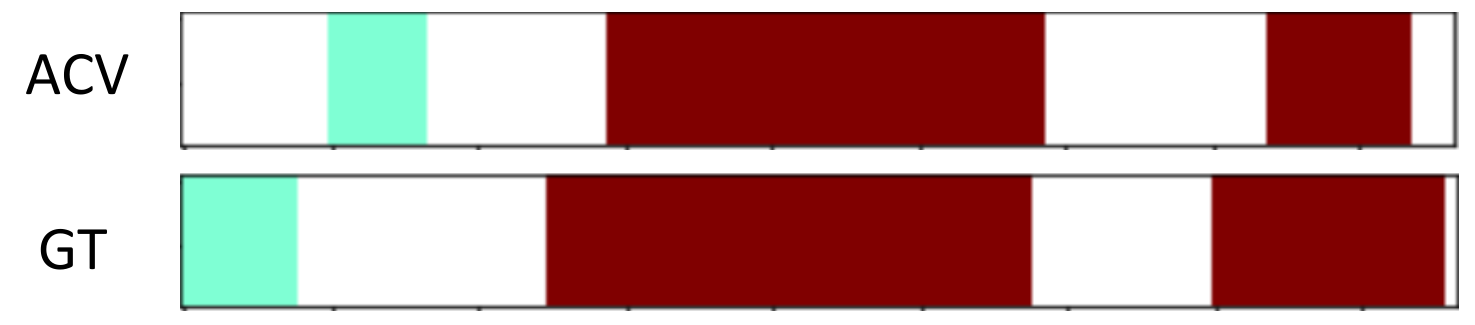}
\end{center}
   \caption{Qualitative result of action alignment on a sample test video \textit{0181} from Hollywood Ext. dataset. Top row: ACV result. Bottom row: ground truth with the color-coded action sequence \{\textcolor{myaquamarine}{OpenCarDoor}, \textcolor{mymaroon}{DriveCar}, \textcolor{mymaroon}{DriveCar}\}. We mark the background frames in white. In general, ACV successfully aligns the actions with video frames, but may incorrectly detect action boundaries.}
\label{fig:align_example}
\end{figure}

In the task of action alignment, we have access to the unordered set of actions $C$ truly present in a test video. Thus, our Monte Carlo sampling can be constrained to uniformly sample only actions in the given ground-truth set $C$. In comparison to the state of the art, Tab.~\ref{Table:alignment} shows that ACV improves action alignment in terms of Mof, Midpoint, and IoD by 4.3\% on Breakfast, 1.1\% on Cooking 2, 5.3\% on Hollywood Ext, respectively. ACV also achieves comparable results to recent approaches that use the stronger transcript-level supervision in both training and action alignment. Fig.~\ref{fig:align_example} shows a qualitative result of action alignment on the test video \textit{0181} from Hollywood Ext. dataset. In general, ACV can successfully align the actions with video frames, but might incorrectly detect action boundaries.

\begin{table}[ht]
\begin{center}
\begin{tabular}{l c c c}
\hline
  & Breakfast & Cooking 2 & Holl.Ext. \\
  Model  & (\textit{Mof}) & (\textit{midpoint}) & (\textit{IoD}) \\
\hline
(Set-supervised) \\
Action Set \cite{richard2018action} & 28.4 & 10.6 & 24.2\\
SCV \cite{li2020set} & 40.8  & 15.1 & 35.5\\
Our ACVnoreg & 43.3  & 15.8 & 38.2\\
Our ACV & {\bf 45.1}  & {\bf 16.2} & {\bf 40.8}\\
\hline
\multicolumn{2}{l}{\text{(Transcript-supervised)} }\\
ECTC \cite{huang2016connectionist} & $\sim$35 & - & $\sim$41\\
HTK \cite{kuehne2017weakly} & 43.9 & - & 42.4\\
OCDC \cite{bojanowski2014weakly} & - & - & 43.9\\
HMM+RNN \cite{richard2017weakly} & - & - & 46.3\\
TCFPN \cite{ding2018weakly} & 53.5 & - & 39.6\\
NN-Viterbi \cite{richard2018neuralnetwork} & - & - & 48.7 \\
D3TW \cite{chang2019d3tw} & 57.0 & - & 50.9 \\
CDFL \cite{Li_2019_ICCV} & 63.0 & - & 52.9 \\
\hline
\end{tabular}
\end{center}
\caption{Evaluation of action alignment that has access to ground-truth action sets of test videos. Our ACV outperforms the state of the art on the three datasets. Also, ACV gives comparable results some transcript-supervised prior work uses stronger supervision in training. The dash means that the result is not reported in the respective paper.}
\label{Table:alignment}
\end{table}

%-------------------------------------------------------------------------

\section{Conclusion}\label{sec:Conclusion}
In this paper, we have addressed set-supervised action segmentation and alignment. Our key contribution is a new Anchor-Constrained Viterbi (ACV) algorithm aimed at generating a frame-wise pseudo-ground truth, which is subsequently used for fully-supervised training of our HMM and MLP models. ACV is an efficient approximation to the NP-hard all-color shortest path problem. Efficiency comes from our estimation of anchor segments that constrain the domain of valid candidate solutions in which we efficiently find the MAP action segmentation. Our approach outperforms state of the art on three benchmark datasets for both action segmentation and alignment, without increasing complexity in training and testing. %Our approach also shows comparable results to some recent methods which have access to stronger transcript-level supervision in training. 
Various ablation studies demonstrate effectiveness of individual components of our model.
\\

\noindent{\bf{Acknowledgement}}. This work was supported in part by DARPA XAI Award N66001-17-2-4029 and DARPA MCS Award N66001-19-2-4035.

{\small
\bibliographystyle{ieee_fullname}
\bibliography{egbib}
}

\end{document}